\documentclass[times,twocolumn,final,authoryear]{elsarticle}
\usepackage{pifont} 
\usepackage{hyperref} 

\usepackage{prletters}
\usepackage{framed,multirow}

\usepackage{amssymb}
\usepackage{latexsym}

\usepackage{url}
\usepackage{xcolor}
\definecolor{newcolor}{rgb}{.8,.349,.1}

\journal{Pattern Recognition Letters}

\begin{document}

\thispagestyle{empty}

\ifpreprint
  \setcounter{page}{1}
\else
  \setcounter{page}{1}
\fi

\begin{frontmatter}

\title{ACDnet: An Action Detection network for real-time edge computing based on flow-guided feature approximation and memory aggregation}

\author{Yu Liu\corref{cor1}} 
\cortext[cor1]{Corresponding author:}
\ead{yu_liu@etu.u-bourgogne.fr}
\author{Fan Yang}
\ead{fanyang@u-bourgogne.fr}
\author{Dominique Ginhac}
\ead{dominique.ginhac@ubfc.fr}
\address{ImViA EA7535, Univ. Bourgogne Franche-Comt\'e, Dijon 21078, France}

\received{1 May 2013}
\finalform{10 May 2013}
\accepted{13 May 2013}
\availableonline{15 May 2013}
\communicated{S. Sarkar}

\begin{abstract}
Interpreting human actions requires understanding the spatial and temporal context of the scenes. State-of-the-art action detectors based on Convolutional Neural Network (CNN) have demonstrated remarkable results by adopting two-stream or 3D CNN architectures. However, these methods typically operate in a non-real-time, offline fashion due to system complexity to reason spatio-temporal information. Consequently, their high computational cost is not compliant with emerging real-world scenarios such as service robots or public surveillance where detection needs to take place at resource-limited edge devices. 
In this paper, we propose ACDnet, a compact action detection network targeting real-time edge computing which addresses both efficiency and accuracy. It intelligently exploits the temporal coherence between successive video frames to approximate their CNN features rather than naively extracting them. It also integrates memory feature aggregation from past video frames to enhance current detection stability, implicitly modeling long temporal cues over time.
Experiments conducted on the public benchmark datasets UCF-24 and JHMDB-21 demonstrate that ACDnet, when integrated with the SSD detector, can robustly achieve detection well above real-time (75 FPS). At the same time, it retains reasonable accuracy (70.92 and 49.53 frame mAP) compared to other top-performing methods using far heavier configurations. Codes will be available at \href{https://github.com/dginhac/ACDnet}{https://github.com/dginhac/ACDnet}.  
\end{abstract}

\end{frontmatter}


\section{Introduction}
In past years, human action detection has been an active area of research driven by numerous applications: autonomous vehicles, video search engines, and human-computer interaction, etc. As it aims not only to recognize actions of interest in a video, but also to localize each of them, action detection poses more challenges when compared to video classification. The task becomes even more difficult in practical applications when detection is to be performed in an online setting and at real-time speed. For instance, time-critical scenarios such as autonomous driving demand instant detection in order for machines to react immediately. Other use cases which seek for mobile or large-scale deployment, such as service robots and distributed unmanned surveillance, require detection or scene meta-data extraction at low-end edge devices. 
In general, edge devices (e.g., embedded systems) have limited computational power and are only compliant with resource-efficient detection algorithms. 

Following the success of Convolutional Neural Network (CNN) in diverse computer vision tasks, modern action detectors are mainly based on CNN. In particular, fast object detectors have been widely adopted to spatially localize action instances at each frame (\cite{singh2017online}, \cite{zhao2019dance}). Naturally, effective temporal modeling plays an imperative role for identifying an action. To reason both spatial and temporal context, \cite{simonyan2014two} pioneered the two-stream CNN framework which aggregates spatial and temporal cues from separate networks and input modalities (RGB and optical flow). Such an approach has motivated many state-of-the-art methods in the field of action recognition and detection. Alternatively, 3D CNN (\cite{carreira2017quo}) which performs spatio-temporal feature learning on stacked frames has also been increasingly explored to tackle video analysis tasks. 

Despite recent advances in action detection, existing methods are inherently sub-optimal in two aspects. First, consecutive video frames exhibit high appearance similarity. Extracting frame features without taking into account this inter-frame similarity introduces redundancy. Moreover, the increased system complexity associated with employing two-stream or 3D CNN models is not proportionally reflected in the detection accuracy. In contrast, the above inevitably raises computational requirements associated with motion extraction and 3D convolution operation, prohibiting practical deployment on edge devices.

This work focuses on action detection solutions more pertinent to the criteria of realistic applications. To address the aforementioned limitations, we first exploit the temporal coherence among nearby video frames to enhance detection efficiency. This is embodied by performing feature approximation at the majority of frames in a video, mitigating re-extraction of similar features from neighboring frames. 
Furthermore, we hypothesize that a less expensive framework can effectively extract meaningful temporal contexts. Here, we adopt a multi-frame feature aggregation module, which recursively accumulates 2D spatial features over time to encapsulate long temporal cues. Such feature aggregation implicitly models temporal variations of actions and facilitates understanding degenerated frames with limited visual cues. 

To the best of our knowledge, this is the first attempt applying feature approximation and aggregation techniques to achieve efficient action detection which can benefit resource-limited devices. To summarize, our contribution is three-fold:

\begin{itemize} 
\item We propose an integrated detection framework, ACDnet, to address both detection efficiency and accuracy. It combines feature approximation and memory aggregation modules, leading to improvement in both aspects. 

\item Our generalized framework allows for smooth integration with state-of-the-art detectors. When incorporated with SSD (single shot detector), ACDnet could reason spatio-temporal context well over real-time, more appealing to resource-constrained devices.

\item We conduct detailed studies in terms of accuracy, efficiency, robustness and qualitative analysis on public action datasets UCF-24 and JHMDB-21.
\end{itemize}

\section{Related work}
Recent advancements in action detection are largely led by building upon successful cases in object detection and action recognition. Here, we briefly review these relevant topics.

\textbf{Object detection} based on CNN methods can be grouped into two families. The two-stage approach such as Faster R-CNN by \cite{ren2015faster} and R-FCN by \cite{dai2016r} first extracts potential object regions from images, on which it performs object classification and bounding box regression on features corresponding to each proposed location. Such a sequential pipeline imposes a bottleneck to real-time inference. 
Alternatively, single-stage detectors such as YOLO proposed by \cite{redmon2017yolo9000}, or SSD in \cite{liu2016ssd}, remove the intermediate region proposal, directly achieving bounding box regression and classification in a single forward-pass. Bypassing the intermediate bottleneck enables real-time detection at the cost of minor accuracy drop.

A number of research focuses on video object detection instead of the image domain. Popular approaches such as \cite{han2016seq} and \cite{kang2017t} exploit videos' temporal consistency by associating detection boxes and scores from multiple frames. Similarly but on the feature level, \cite{hetang2017impression} and \cite{zhu2017flow} aggregate multiple frame features to enhance detection accuracy. On the other hand, \cite{zhu2017deep} leverage the temporal redundancy among video frames to improve detection efficiency. Their framework propagates features from a sparse set of key frames to successive ones by motion to avoid re-extracting similar object features. In a similar spirit, \cite{liu2018mobile} propagate frame-level information across frames using a recurrent-convolutional architecture. 
%

\textbf{Action recognition} is typically treated as a classification task on trimmed videos (\cite{yao2019review}). In addition to spatial features, reasoning temporal information across multiple frames is also crucial. Among different temporal modeling techniques, the two-stream architecture in \cite{simonyan2014two} demonstrates state-of-art performance. Its framework consists of two feed-forward pathways, with one CNN learning spatial features from RGB stream and the other one learning motion features from optical flow stream. The two streams are trained and run inference independently to aggregate complemented features (\cite{feichtenhofer2016convolutional}). Even though such a framework can exploit existing 2D CNN backbones, fine-grained optical flow is expensive to extract. Thus, flow images are typically pre-computed, which do not conform to the online workflow demanded in real-world scenarios.

Recently, 3D CNNs have been increasingly explored (\cite{carreira2017quo}, \cite{li2019large}) along with the release of large-scale action dataset Kinetics. They utilize 3D kernels to jointly perform spatio-temporal feature learning from stacked RGB frames, achieving comparable and even superior modeling capability than two-stream CNN. However, these models inherently suffer from higher number of parameters and computational cost than their 2D counterparts, making their deployment on resource-constrained devices impractical. 


\textbf{Efficient spatio-temporal modeling.} To alleviate the high computational cost associated with flow extraction, several studies seek alternative motion representations that are easier to compute. These include feature-level displacement (\cite{jiang2019stm}, \cite{sun2018optical}), or simply taking the RGB difference between adjacent frames (\cite{wang2016temporal}). On the other hand, to reduce the complexity of 3D CNN, decoupled architectures such as P3D (\cite{qiu2017learning}) and R(2+1)D (\cite{tran2018closer}) have been studied. 
Alternatively, TSM proposed by \cite{lin2019tsm} handles temporal convolution as channel-shifting operators to fuse spatial features from different time steps. Their approach has demonstrated effectiveness on edge devices such as Jetson Nano and Galaxy Note8.

\textbf{Spatio-temporal action detection} simultaneously addresses action localization and classification in time and space. Leading approaches often leverage CNN object detectors as the core building block. The extension mainly consists of adopting the two-stream framework, fusing complementary detection results from both spatial and temporal stream to acquire frame-level detection, as demonstrated in \cite{singh2017online}. For temporal localization, detection at each frame is then linked over time to construct action tubes (\cite{peng2016multi}). 
Beyond detection at the frame-level, \cite{kalogeiton2017action} and \cite{li2020actions} adopt a clip-based approach, which exploits stacking multiple frame features to capture temporal cues on top of the two-stream architecture. In this case, actions are regressed and inferred directly on action cuboids. 

Inspired by the latest adoption of 3D CNN in action recognition, more recent studies incorporate 3D CNN as the backbone (\cite{yang2019step}, \cite{sun2018actor}, \cite{girdhar2019video}, \cite{gu2018ava}, \cite{wei2019p3d}). In addition, various ways of fusing spatial and temporal context have also been investigated. Besides aggregating at the detection level (e.g., union of detection results), others perform feature-level fusion. These include the use of $1\times1$ convolution (\cite{ali2018real}), attention model (\cite{kopuklu2019you}) or conditional normalization (\cite{zhao2019dance}). Such approaches allow a part of CNN layers to adaptively learn from the fused features. 
\section{ACDnet}

\begin{figure*}
  \centering
  \includegraphics[width=5.2 in]{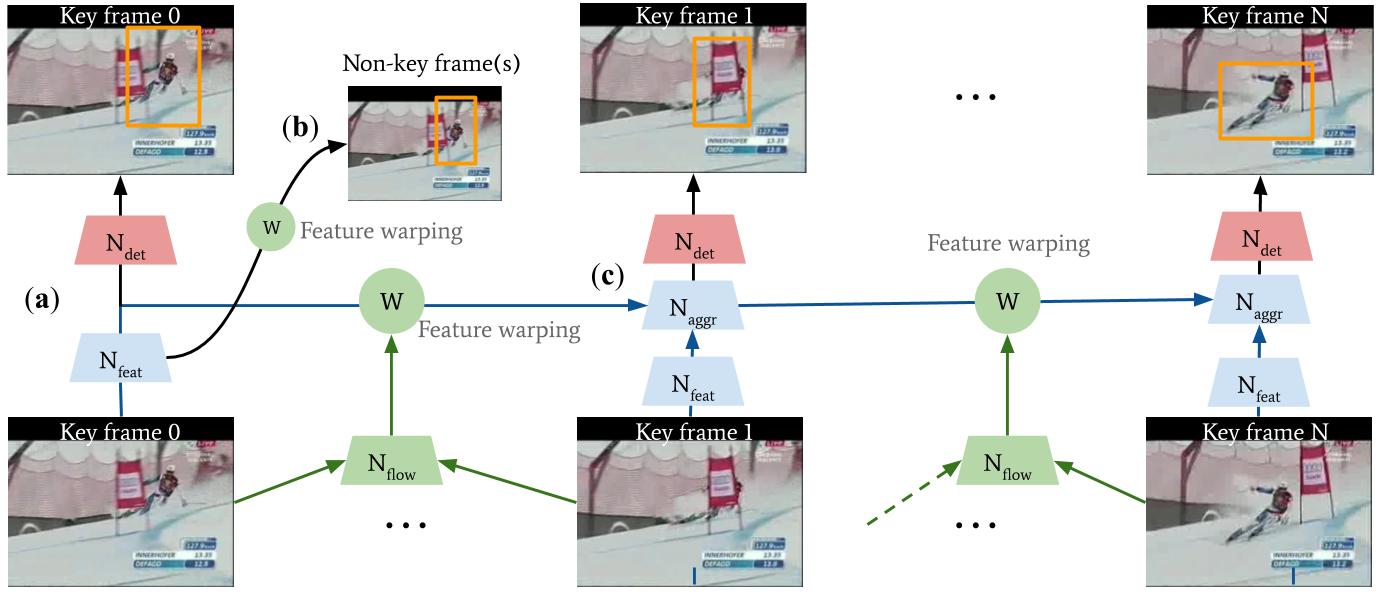}
  \caption{Illustration of ACDnet inference pipeline. (a) At the \textbf{initial frame}, features are obtained from the feature extraction sub-network ($N_{feat}$). (b) For \textbf{non-key frames} (dense), the flow sub-network ($N_{flow}$) estimates a pair of flow field and position-wise scale map between the non-key frame and its preceding key frame. The resulted flow field is used to propagate appearance feature, which is then refined by the scale map via element-wise multiplication. (c) At \textbf{key frames} (sparse), new features are extracted. They are then aggregated with those from the past key frames (memory features) via $N_{flow}$ and the aggregation sub-network ($N_{aggr}$). The fused features will be used for detection ($N_{det}$) and also passed along as the updated memory.}
    \label{fig:overall_method_pipeline}
\end{figure*}

Our objective is to perform detection in an online manner for every incoming frame of a video. The proposed ACDnet which consists of the feature approximation and aggregation module, is summarized in Figure \ref{fig:overall_method_pipeline}. 


\subsection{Feature approximation by motion guidance}
Video content varies slowly over consecutive frames. This phenomenon is more so reflected in the corresponding CNN feature maps which capture high-level semantics. 
Intuitively, the shared appearances among neighboring frames can help to propagate essential information for a given task. The practice of propagation in \cite{zhu2017deep} has established success to enhance object detection efficiency in videos, which motivates our feature approximation module.  

Within the approximation scheme, the heavier feature extraction sub-network, $N_{feat}$, only operates on a sparse set of key frames during inference. The features of successive non-key frames are obtained by spatially transforming those from their preceding key frames via two-channel flow fields. The workflow can be summarized by the following equations. Let $M_{i \to k}$ be the two-channel flow field capturing relative motion  from the current frame $I_i$ to its previous key frame $I_k$ (horizontal and vertical direction). Then, feature approximation (also referred as feature propagation) is realized according to inverse warping:  

\begin{equation}
F_i = W(F_k, M_{i \to k})
\label{eq:warping_global}
\end{equation}
where $F_k$ is the key frame feature, and $F_i$ is the newly warped feature corresponding to $I_i$. $W$ denotes the inverse warping operation to sample the correct key frame features and assign them to the warped ones. Inverse warping is necessary to ensure every location $p$ at the warped feature can be projected back to a point $p + \Delta p$ at the key frame feature, where $\Delta p = M_{i \to k} (p)$. Concretely, the warping operation $W$ is performed as:

\begin{equation}
f_i^c (p_i) = \sum_{p_{k}} G(p_{k}, p_i + \Delta p) f_k^c(p_k) 
\label{eq:warping_detail}
\end{equation}

In Equation \ref{eq:warping_detail}, $f_i^c$ and $f_k^c$ denote the $c^{th}$ channel of feature $F_i$ and $F_k$, respectively; $G$ denotes the bilinear interpolation kernel. Every location $p_i$ in the warped feature map undergoes this warping scheme to sample features from key frames, independently for each feature channel $c$. The warping operation is much lighter compared to layers of convolution for feature extraction. Consequently, by applying feature approximation on a dense set of non-key frames, computation is greatly reduced. 


Previous methods on action-based tasks typically acquire motion features from accurate optical flow using non-learning-based algorithms. However, computing flows in such a way imposes a bottleneck to real-time and online detection due to high consumption of time or requiring to pre-compute flow results. In contrast, ACDnet integrates a fast flow estimation sub-network, $N_{flow}$, to predict flow fields. In our case, optical flow serves to spatially transform CNN features; it does not need to capture fine-grained motion details and has the same height and width as the corresponding feature. Using such a learning-based flow estimator also allows it to be jointly trained with all other sub-networks specific to the task of action detection. 

In detail, the flow sub-network take ($I_k$, $I_i$) as input, and generates a pair of motion field and position-wise scale map. Given that $H$, $W$, and $C$ denote height, width and channel of $F_k$, then the flow field $M_{i \to k}$ is of size $H \times W \times 2$ , and the scale map is $H \times W \times C$, whose dimension matches that of $F_k$ to be warped. After the inverse warping described by Equation \ref{eq:warping_global}, the warped feature $F_i$ is refined by multiplying the scale map in an element-wise way. Any $F_k$ and $F_i$ would be fed to the shared detection sub-network, $N_{det}$, to obtain final detection. This workflow is illustrated in Figure \ref{fig:overall_method_pipeline} (a) and (b).





\subsection{Memory feature aggregation}
Propagating features across frames reduces the computation cost associated with feature extraction. However, since most features are now approximated and heavily dependent of the quality of the precedent key frame features, we adopt a memory aggregation module as inspired by \cite{hetang2017impression} to enhance the feature representation at key frames. Given incoming video frames, the core of memory aggregation is to reinforce features of a target frame by recursively incorporating supportive and discriminating context from the past. This allows implicit spatio-temporal modeling without explicitly extracting motion features. In addition, in cases when the current frame is deteriorated, an action can still be inferred with the supportive visual cues from memory. Figure \ref{fig:overall_method_pipeline} (c) gives an example when such memory aggregation could be useful. 

Memory aggregation shares the same warping operation used for feature approximation. ACDnet takes a sparse and recursive approach to aggregate memory features only at key frames, due to similar appearances shared among nearby frames. Given two succeeding key frames $I_{k1}$ and $I_{k2}$, where $I_{k2}$ is the more recent one in time, memory aggregation follows Equation \ref{eq: memory_aggregation}:

\begin{equation}
F_{k2\_aggregated} = w_{k1} \otimes F_{k1}^\prime + w_{k2} \otimes F_{k2}
\label{eq: memory_aggregation}
\end{equation}
where $F_{k1}^\prime = W(F_{k1}, M_{k2 \to k1})$, is the warped feature of $I_{k1}$ to spatially align its position with that of $I_{k2}$. The position-wise weights $w_{k1}$ and $w_{k2}$ both have the same height and width as $F_{k1}^\prime$ and $F_{k2}$. These weights are normalized and determine the importance of memory feature at each location $p$ with respect to the target frame feature map ($w_{k1}(p) + w_{k2}(p) = 1$). All channels of a feature share the same spatial weights. 

The weights $w_{k1}$ and $w_{k2}$ are adaptively calculated based on the similarity of memory and target features. We estimate feature similarity by first projecting them into an embedding space via a few convolution layers, and then computing the cosine similarity between the embedded features. Finally at the current key frame, the weighted sum of the memory and current features will be fed to the detection sub-network and passed along as the new memory. 


\subsection{Training procedure}
\begin{figure}
  \centering
  \includegraphics[width=3.0in]{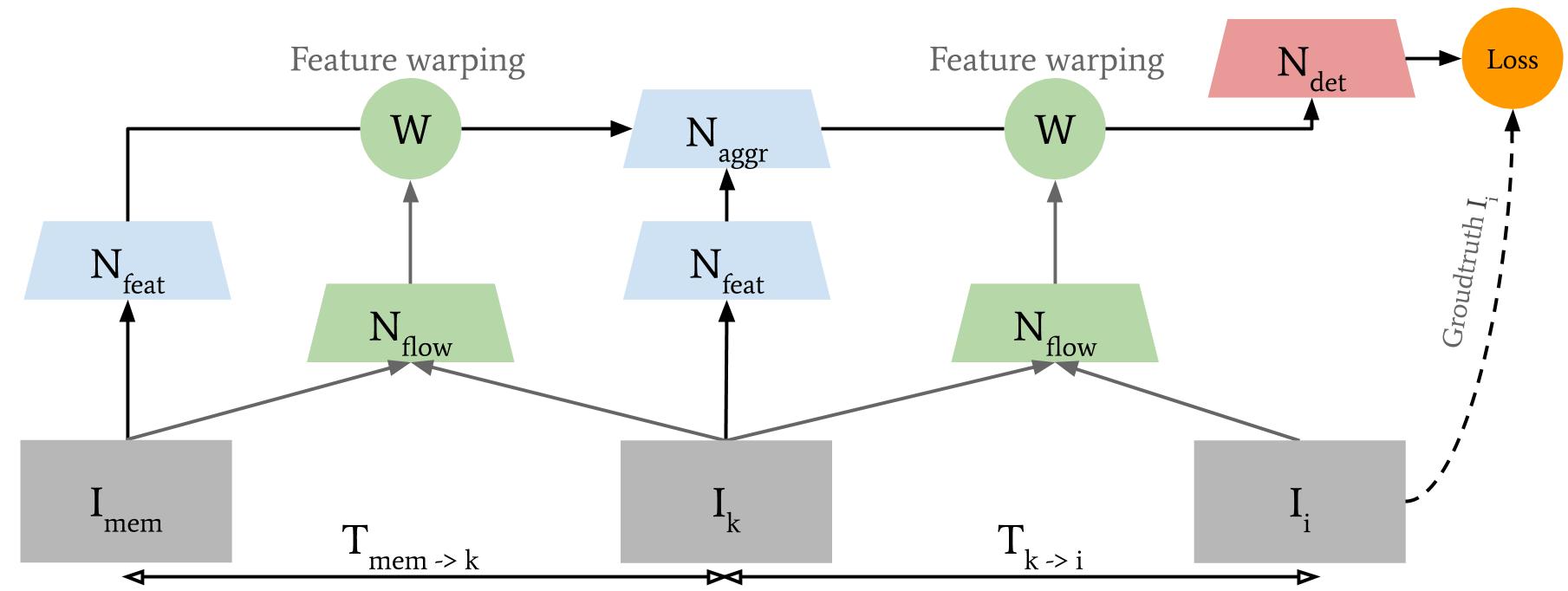}
  \caption{Training procedure. Each mini-batch consists of three frames ($I_{mem}$, $I_k$, and $I_i$) and the groudtruth of $I_i$.}
    \label{fig:training_procedure}
\end{figure}

ACDnet follows a three-frame training scheme, as depicted in Figure \ref{fig:training_procedure}. From each training mini-batch, frame $I_i$ and two precedent video frames ($I_k$ and $I_{mem}$) are selected, whose features simulate key frame and memory features respectively. The offset between $I_i$ and $I_k$ is a random number from 0 to $T_{k\to i}$, and the offset between $I_k$ and $I_{mem}$ is fixed at $T_{mem\to k}$. 

The feature maps $F_{mem}$ and $F_k$ are first extracted from $I_{mem}$ and $I_k$ respectively. Two sets of flow fields, namely, the relative motion between $I_k$-$I_{mem}$, and $I_i$-$I_{k}$ are also estimated. The former flow is used to propagate $F_{mem}$ to $F_k$ to simulate the occurrence of memory feature aggregation following Equation \ref{eq: memory_aggregation}. Finally, the fused feature is warped with the second flow (simulating feature approximation) following Equation \ref{eq:warping_global}, which will be the final feature map for $N_{det}$. Under this training mode, only the groundtruth of $I_i$ is needed to determine losses, which are back-propagated to update all sub-networks.  

\subsection{Adaptation for multi-feature scale detector}
Workflows of feature approximation and aggregation are generic for video-based tasks. ACDnet further employs SSD, an one-stage detector to fulfill the objective of high-speed action detection potentially for embedded vision systems. In particular, the SSD300 model is chosen due to its superior speed.


In SSD, a set of auxiliary convolutional layers are progressively added after the base network (e.g., VGG16 in a standard SSD) to extract features at multiple scales. This creates multiple feature maps where the detector makes prediction for objects of various sizes. Consequently, adopting the described framework in SSD requires feature approximation and memory aggregation to be handled for features at all scales.

\begin{figure}[!t]
  \centering
  \includegraphics[width=3.0in]{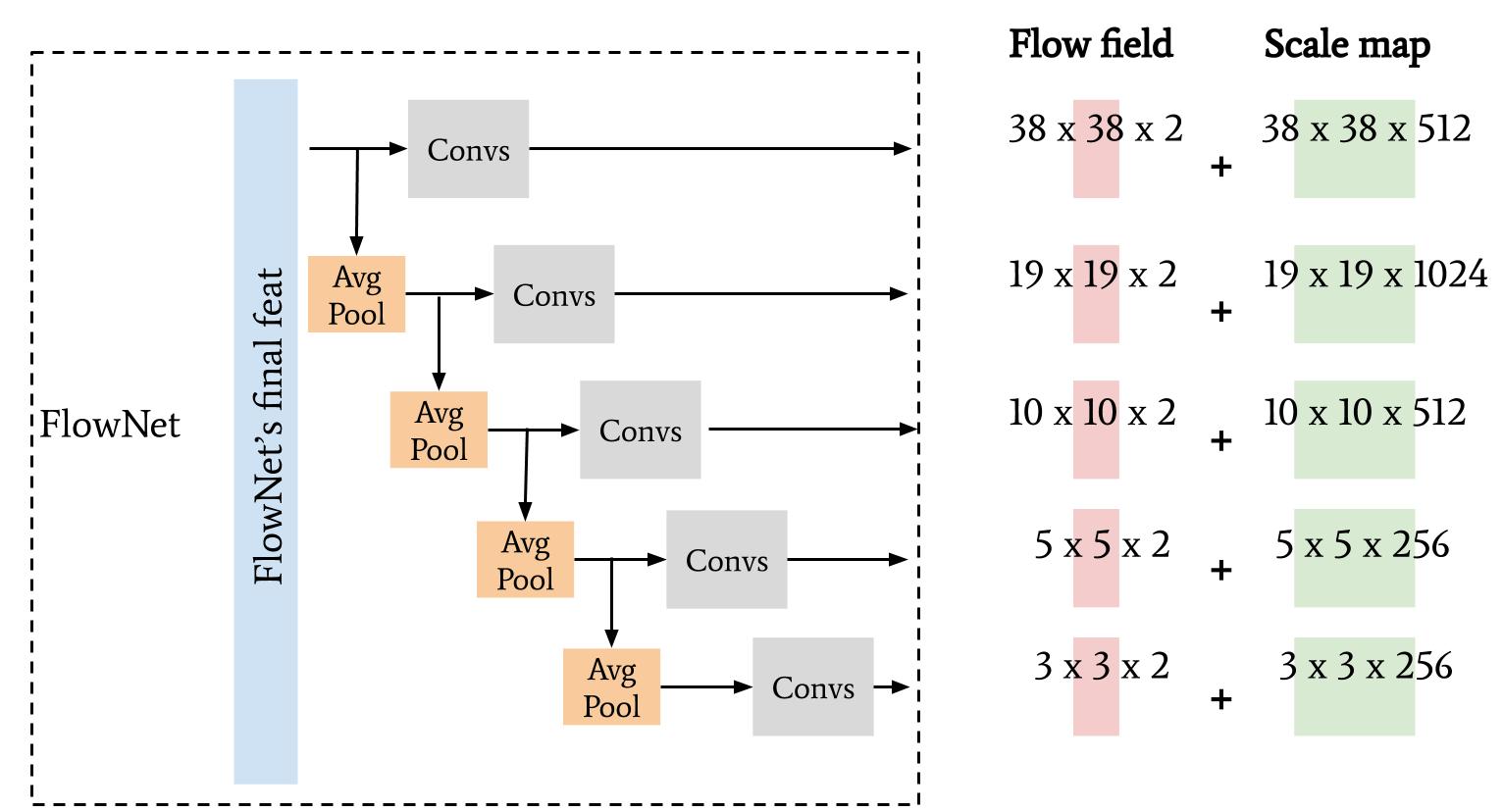}
  \caption{Flow estimation sub-network adapted for multi-scale feature approximation and aggregation. The depicted design corresponds to the architecture of SSD300 and FlowNet.}
    \label{fig:flownet_output}
\end{figure}

To enable multi-level feature approximation, we duplicate $N_{flow}$'s flow prediction layer into several branches. The number of branches matches that of the feature maps; the branches' outputs are also progressively resized via average pooling according to sizes of SSD's feature maps. Then, each branch reconstructs a pair of flow field and scale map in accordance with the dimension of SSD feature (refer to Figure \ref{fig:flownet_output}). To cope with multi-level feature approximation and aggregation, Equation \ref{eq:warping_global} and \ref{eq: memory_aggregation} are also generalized to take place at each feature level independently. Note that the standard SSD300 applies detection at six feature scales. Nevertheless, we only use the first five of them, as the dimension of the last feature map becomes a 1D vector resulting from progressive resizing, which is not feasible for feature approximation governed by 2D spatial warping. 

\section{Experimental results}
\subsection{Experimental Setup}
\textbf{Dataset}. Our proposed methods are evaluated on two popular action datasets: \textbf{UCF-24} and \textbf{JHMDB-21}. The former one released by \cite{soomro2012ucf101} is composed of 3207 sports videos of 24 action classes. Following previous work, we use 2290 of these video clips for training.  
The latter collected by \cite{jhuang2013towards} consists of 928 short videos divided into three splits, with 21 action categories in daily life. Each video is trimmed and has a single action instance. We report our experimental results on the average of three splits for this dataset. 

\textbf{Network architectures}.
ACDnet incorporates the following sub-networks: SSD300 (with VGG16 backbone), FlowNet (\cite{dosovitskiy2015flownet}) and feature embedding. Feature embedding contains five branches for measuring feature similarity at five different scales. Each embedding branch has a bottleneck design of three $1\times1$ convolution layers interleaved with ReLU non-linearity, where the number of filters corresponds to the number of channels at each feature level $l$: $feat^l_{channel}/2$, $feat^l_{channel}/2$ and $feat^l_{channel}\times2$, respectively. 

FlowNet is modified to also generate five sets of flow fields and position-wise scale maps, each pair being used for warping and refining designated features. We initialize the weights of the first two branches of flow generation layers using FlowNet's pre-trained weights. Considering the later three flow outputs are spatially much smaller than that of the original FlowNet, we randomly initialize the weights of those branches.



\textbf{Training}.
Images are resized to $300\times300$ for training and inference. Training is conducted by stochastic gradient descent. To address data imbalance among different actions, from each training video clip of UCF-24, 15 frames spanning the whole video are evenly sampled as the training set. Since video clips of JHMDB-21 are generally short ($\leq$ 40), we evenly sample 10 frames from each clip for training. Specifically, both $T_{mem \to k}$ and $T_{k \to i}$ are set to 10 during training. These chosen values correspond to the key frame interval used during inference, which is also fixed at 10 in our experiments unless specified.

We apply different hyperparameters on the two datasets. UCF-24 is trained for $100K$ iterations; the learning rate is initialized as 0.0005 and reduced by a factor of 0.1 after the $80K^{th}$ and $90K^{th}$ iterations. Weights of VGG16's first two convolution blocks are frozen. For JHMDB-21, due to its smaller training and testing size, we observe that detection accuracy tends to fluctuate significantly between successive epochs. Hence, we empirically train this dataset for $20K$ iterations with learning rate initialized as 0.0004 and reduced by a factor of 0.5 after the $8K^{th}$ and $16K^{th}$ iteration. During its training, the first three convolution blocks of VGG16 are frozen. In addition, all layers of FlowNet until the flow generation layers (the five branches at the end of our modified model) are also frozen to further reduce the risk of overfitting.

All sub-networks are trained jointly (also evaluated) on an NVIDIA Quadro P6000 GPU using a training batch size of 8. For the rest of hyperparameters and data augmentation methods, we follow the same setup as the original SSD by \cite{liu2016ssd}. The weights of VGG16 and FlowNet are pre-trained using ImageNet and the Flying Chair dataset respectively.   



\subsection{Ablation study}
Our proposed architecture has been evaluated in terms of accuracy, efficiency and robustness over several network configurations. The standard frame-level mean average precision (F-mAP) and frame-per-second (FPS) have been used as the evaluation metrics. Specifically, FPS is measured based on the complete detection pipeline, including data loading and model inference using batch size of 1.The Intersection-over-Union threshold is set to 0.5 throughout all experiments. For brevity, we refer to feature approximation and memory feature aggregation as \textbf{FA} and \textbf{MA} respectively when presenting their results.  

\begin{figure}[h]
    \centering
    \includegraphics[width=2.8 in]{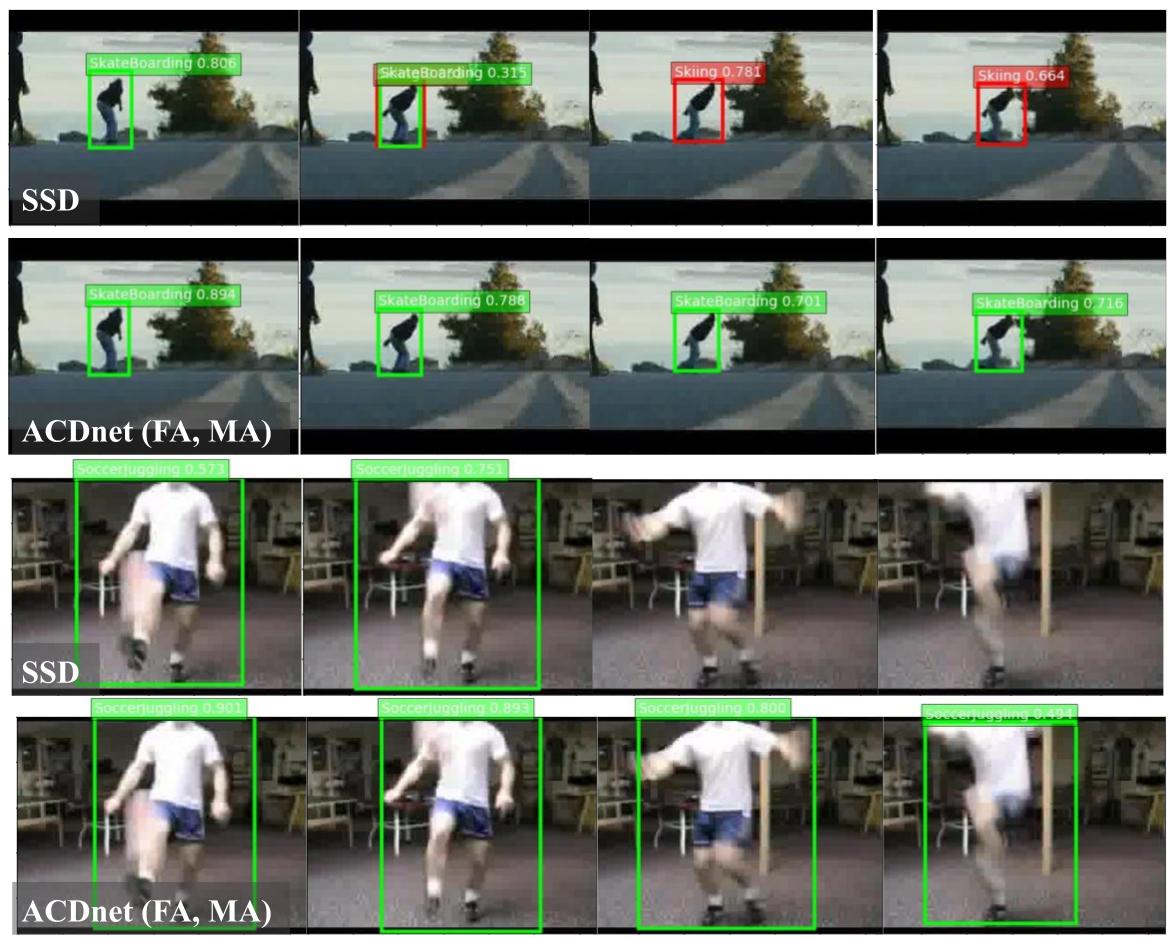}
    \caption{Examples where ACDnet (FA, MA) improves the baseline SSD. Green / Red boxes correspond to correct / incorrect detection respectively. 
    }
    \label{fig:qualitative_examples}
\end{figure}

\textbf{Accuracy.} F-mAP results of different configurations are reported in Table \ref{tab:accuracy_results}. From both datasets, we observe a decrease of accuracy when only feature approximation is included. However, the accuracy drop can be compensated by the addition of memory aggregation, which exceeds the accuracy of the stand-alone SSD. Figure \ref{fig:qualitative_examples} shows some examples of how the memory aggregation module benefits detection. Overall, we remark that aggregating multipleframe features over time, even in a sparse manner, improves models' abilities to more confidently discriminate among different actions.

{
\begin{table}[!t]
\caption{F-mAP results for different configurations. 
}
\centering
  \begin{tabular}{l|c|c|c|c|c}
\hline
ACDnet
\\ \hline
                                             
SSD & \ding{51} & \ding{51} & \ding{51} & \ding{51} & \ding{51}  \\
FA  & & \ding{51} & \ding{51} & \ding{51} & \ding{51} \\
Scale map &  &  & \ding{51} &  & \ding{51}   \\
MA &  &  &  & \ding{51} & \ding{51}   \\
\hline
F-mAP
\\ \hline
UCF-24 & 67.32 & 65.84 & 67.23 & 68.06 & \textbf{70.92}   \\
JHMDB-21 & 47.90 & 46.65 & 46.69 & 49.37 & \textbf{49.53}  \\
\hline
\end{tabular}%

\label{tab:accuracy_results}
\end{table}
}

\begin{figure}
  \centering
  \includegraphics[width=3.0in]{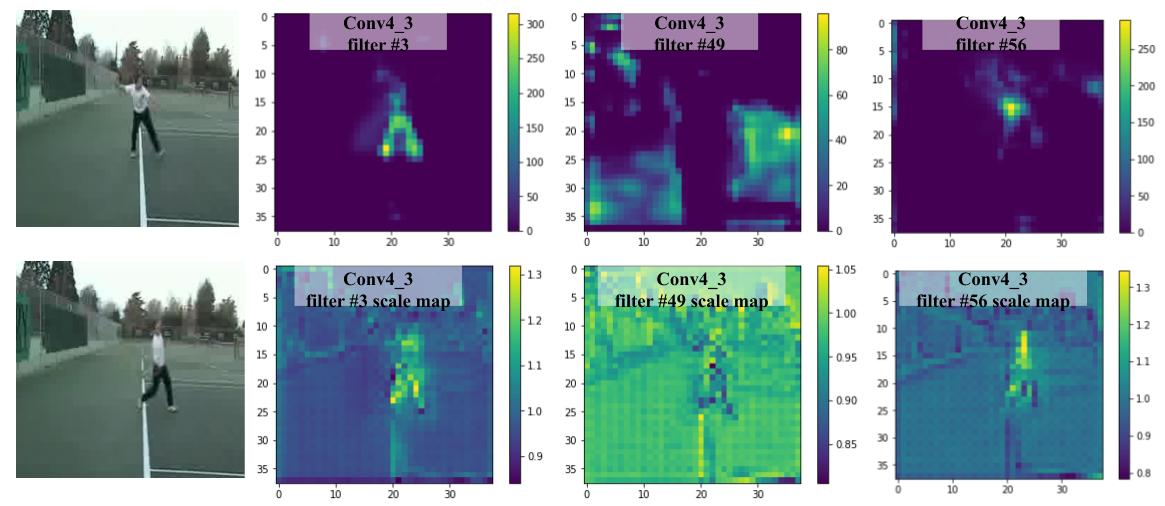}
  \caption{Position-wise scale maps produced by our modified FlowNet. The scale map (bottom row) only reinforces activation (top row) associated with the actor by up-scaling, without altering activation in other feature regions.}
    \label{fig:flow_scale_maps}
\end{figure}

We also examine the effect of separate branches of position-wise scale maps designed for refining visual features. Our results indicate that such refinement mildly improves detection accuracy. The scale maps serve as implicit attention maps which reinforce feature responses associated with moving actors (elaborated in Figure \ref{fig:flow_scale_maps}).  

Even though similar result patterns can be seen from both datasets, the benefit of memory aggregation appears less prominent in JHMDB-21. This could result from the fact that each video clip in JHMDB-21 is much shorter (40 frames or less). As MA is performed sparsely at every $10^{th}$ frame, its impact is limited to 2-3 aggregation per clip. Furthermore, we observe that motions in several JHMDB-21 clips are relatively small. In these clips, key frames far apart still appear fairly identical, limiting additional visual cues to be propagated.  

\textbf{Efficiency} is evaluated on UCF-24 by simultaneously inspecting accuracy, run time and number of parameters of various configurations. Here, we assume the use of scale map refinement if applicable. As shown in Table \ref{tab:speed_results}, ACDnet (SSD, FA, MA) outperforms the stand-alone SSD in both speed and accuracy. This suggests it is relevant to handle inter-frame redundancy, and that long-range memory fusion is effective for collecting more discriminating features. Regarding the number of required parameters, the increase in ACDnet (SSD, FA) from stand-alone SSD is associated with the addition of FlowNet, which can be replaced by much lighter architectures in the future. Likewise, the increase with the addition of MA module corresponds to the extra embedding layers for measuring feature similarity at various scales. In terms of run time, the speed drop with MA is incurred by the additional operations at key frames (except for the first one), where flow estimation, feature extraction, similarity measure and aggregation all take place. 

To examine how our generic architecture performs on a different detection framework, we conduct the same experiments while incorporating ACDnet with R-FCN, a state-of-the-art two-stage detector. The run time improvement brought by feature approximation is more significant with R-FCN, due to it using a much deeper backbone for feature extraction. The number of additional parameter needed to carry out memory aggregation is less too for R-FCN, as it is designed to perform prediction on a single-scale feature (needing only one branch for the embedding and flow sub-networks). 
Overall, when taking into account run time, memory consumption and obtained accuracy together, our results still strongly favor the SSD-based ACDnet. 

{
\begin{table}[!t]
\caption{Performance of different configurations on UCF-24. 
}
\centering
  \begin{tabular}{lccc}
\hline
                                               & F-mAP & FPS & \# params \\ \hline
SSD  & 67.32 & 70 & 26.8M \\
ACDnet (SSD, FA)    & 67.23 & 85 & 50.8M \\
ACDnet (SSD, FA, MA) & 70.92 & 75 & 57M \\ \hline
R-FCN  & 68.2 & 15 & 60M \\ 
ACDnet (R-FCN, FA) & 66.19 & 33.5 & 85.7M \\
ACDnet (R-FCN, FA, MA) & 68.31 & 32 & 89.6M \\
 \hline
\end{tabular}%

\label{tab:speed_results}
\end{table}
}

\begin{figure}[!t]
  \centering
  \includegraphics[width=3.0in]{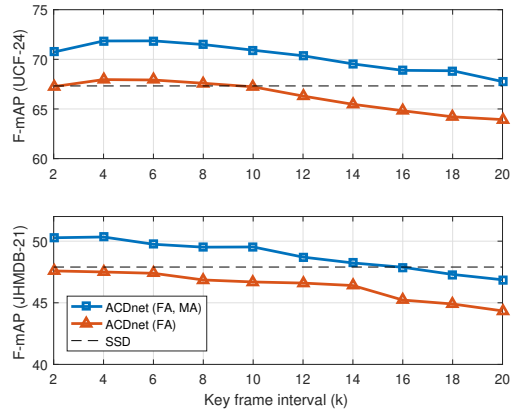}
  \caption{F-mAP under varied key frame intervals.}
    \label{fig:robustness_fmap}
\end{figure}

\textbf{Robustness.} Concerning the robustness of our models trained with a fixed duration $T_{mem \to k}$ and $T_{k \to i}$ (at 10), we evaluate their performances under various key frame intervals ($k$) during inference. Figure \ref{fig:robustness_fmap} displays F-mAP results on both datasets, with $k$ ranging from 2 to 20. Both models (with and without MA) express an overall steady drop in accuracy on the two datasets as $k$ increases. This is reasonable, as the ability of flow fields to correctly encode pixel correspondence diminishes under large motions. However, even when $k$ is large, ACDnet with MA still retains decent accuracy which outperforms or is comparable with the best cases of the other configurations. 

\begin{figure}[!t]
  \centering
  \includegraphics[width=3.0in]{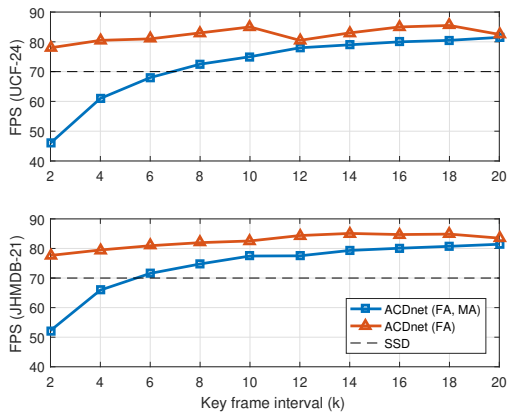}
  \caption{FPS under varied key frame intervals.} 
    \label{fig:robustness_fps}
\end{figure}

Run time is also inspected under the same setting, as shown in Figure \ref{fig:robustness_fps}. It can be observed that ACDnet (FA, MA) exceeds the speed of SSD starting around $k=8$, while the FA-only model is consistently faster. Larger key frame intervals intuitively should lead to further speed boost, as higher ratio of features are approximated. Interestingly, we observe that this pattern is neatly presented when $k \leq 10$. After that, the run time of the examined models begin to saturate. This phenomenon is associated with two factors. On the one hand, as key frame interval increases, the ratio between approximated and real features become less significant. On the other hand, larger key frame intervals introduce more motion which could compromise the quality of approximated features. This results in an increase of low-confident predictions, which take longer for SSD's non-maximum suppression to filter.



\subsection{Comparison with state-of-the-art}


We compare the complete ACDnet (with FA and MA) against state-of-the-art methods in Table \ref{tab:state_of_the_art_comparison}. Since our proposed framework targets lightweight action inference for realistic deployment rather than solely obtaining superior accuracy, only top-performing works which take into account both accuracy and run time are considered for fair comparison.  With this in mind, recent research such as the works of \cite{wei2019p3d} and \cite{gu2018ava} demonstrate impressive accuracy but are excluded from our comparison, as they utilize heavier configurations and omit speed analysis. Alongside performances, comprehensive summary of each method's backbone is also reported for clearer comparison. It should be noted that methods such as ACT, STEP and MOC perform clip-based detection. In other words, they take clips of multiple RGB frames with the support of stacked flow images at once (e.g., five flow for each RGB), predicting action tubelets spanning these RGB frames. In contrast, methods such as YOWO gather supportive contextual cues from multiple frames to augment the target one. These particular attributes are summarized Table \ref{tab:state_of_the_art_comparison} column 4. 

As shown in Table \ref{tab:state_of_the_art_comparison}, ACDnet outperforms the others in terms of run time. This is ascribed to the feature approximation module and our less complex architecture overall. The other methods either adopt two-stream or 3D CNN architectures to capture complemented spatial and temporal features, which raise computation. In addition, preparation of accurate flow using Brox (\cite{brox2004high}) or FlowNet2 (\cite{ilg2017flownet}) is particularly expensive; as a result, all methods employing a second flow stream do not take into account optical flow acquisition when measuring run time (except ROAD using a fast flow estimator by \cite{kroeger2016fast}). In contrast, flow generation in ACDnet is fast and can be carried out in an online setting as it does not aim to encode fine-grained motion features. 

\begin{figure}[!t]
  \centering
  \includegraphics[width=3.2 in]{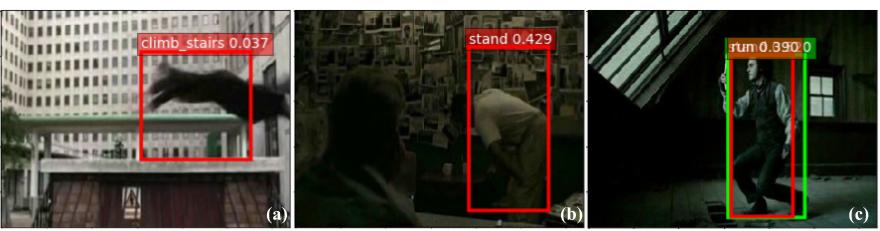}
  \caption{Examples of false detection in JHMDB-21. (a) Correct action: \emph{Jump}. (b) Correct action: \emph{Sit}. (c). Correct action: \emph{Stand}; ACDnet incorrectly predicts two actions (\emph{Stand} and \emph{Run}).}
    \label{fig:fail_examples_jhmdb21}
\end{figure}

In terms of accuracy, ACDnet retains competitive performance on UCF-24. On the other hand, its performance on JHMDB-21 is less impressive compared to the other methods. As opposed to UCF-24, whose classes of sports activities are visually more distinctive, we observe that JHMDB-21 contains more  classes that share similar visual context (for example, \textit{Sit} v.s. \textit{Stand}, and \textit{Run} v.s. \textit{Walk}, etc.). Figure \ref{fig:fail_examples_jhmdb21} demonstrates a few falsely detected examples by our model which result in lower F-mAP in JHMDB-21. Such ambiguous visual context is challenging even for human to confidently infer the correct action unless viewing consecutive frames at once. As shown in column 4 of Table \ref{tab:state_of_the_art_comparison}, ACDnet applies detection on frames far fewer than other methods, which limits its ability to model detailed variations of visual cues over time. In addition, 
JHMDB-21 consists of short clips for which sparse memory aggregation can only take place minimally. The above factors result in ACDnet's less satisfactory accuracy on JHMDB-21. This visual ambiguity could generally be mitigated when examining more frames at once, as demonstrated by all clip-based methods.

Similarly, 3D-CNN-based method such as YOWO also proves effective to learn spatio-temporal features when taking 16 consecutive frames. However, such an approach inevitably raises the computation time; not only from model inference, but also data loading, which is excluded in their reported speed performance. Furthermore, ACDnet achieves comparable accuracy when YOWO employs lighter 3D CNN variants, implying the necessity of deeper models to effectively reason temporal context. In conclusion, our experimental results verify ACDnet's competitive capability to efficiently infer actions with strong visual cues, but its sparse spatio-temporal modeling scheme does not capture temporal cues as effectively as the more expensive two-stream / 3D CNN. 
On the other hand, ACDnet is compact and can achieve inference speed far beyond real-time requirement. This not only permits more seamless deployment potentially on resource-constrained devices, but also can afford to further adopt a clip-based framework or light-weight 3D CNN to improve its accuracy.  


{
\begin{table*}[t]
\caption{State-of-the-art comparison. *For any key frame, ACDnet (FA, MA) fuses the accumulated key frame feature from the past with the current one (considered 2RGB implicitly). For any non-key frame, its feature is approximated based on the preceding key frame feature (considered 1RGB).  
}
\begin{tabular}{l|c|c|c|c|c|c}
\hline
Method  & +2-stream Flow & +3D CNN & \verb|#|frames:\verb|#|det. & UCF24 & JHMDB21 & FPS \\ \hline


ROAD, \cite{singh2017online} & Kroeger & \ding{55} & (1RGB+1FL):1 & 65.66 & -- & 28 \\

ACT, \cite{kalogeiton2017action} & Brox & \ding{55} & (6RGB+30FL):6 & 69.5 & 65.7 & 25 \\

TS-YOLO, \cite{ali2018real} & FlowNet2-SD & \ding{55} & (1RGB+1FL):1 & 71.67 & -- & 25 \\

STEP, \cite{yang2019step} & Brox & $3\times$ 3D conv & (6RGB+30FL):6 & 75 & -- & 21 \\
YOWO(a), \cite{kopuklu2019you} & \ding{55} & 3D-ResNet101 & 16RGB:1 & 80.4 & 74.4 & 34\\
YOWO(b) & \ding{55} & 3DShuffleNetV2 & 16RGB:1 & 71.4 & 55.3 & --\\
YOWO(c) & \ding{55} & 3DMobileNetV2 & 16RGB:1 & 66.6 & 52.5 & --\\
MOC, \cite{li2020actions} & Brox & \ding{55} & (7RGB+35FL):7 & 78 & 70.8 & 25 \\
\hline
ACDnet (FA, MA) & \ding{55} & \ding{55} & $2(1)^{*}$RGB:1 & 70.92 & 49.53 & 75 \\
\hline
\end{tabular}%

\label{tab:state_of_the_art_comparison}
\end{table*}
}
\section{Conclusions and Future works}
In this paper, we present ACDnet, a compact action detection network with real-time capability. By exploiting temporal coherence among video frames, it utilizes feature approximation on frames with similar visual appearances, which significantly improves detection efficiency. Additionally, a memory aggregation module is introduced to fuse multi-frame features, enhancing detection stability and accuracy. The combination of the two modules and SSD detector implicitly reasons temporal context in an inexpensive manner. ACDnet demonstrates real-time detection (up to 75 FPS) on public benchmarks while retaining decent accuracy against other best performers at far less complex settings, making it more appealing to edge device deployment in practical applications. Our future works include further investigation in cost-effective architectures for spatio-temporal modeling and performing temporal localization. For a fully integrated and  resource-efficient vision system, lightweight alternatives of the current sub-networks will be explored, and we will precisely customize candidate solutions for embedding them on edge devices such as NVIDIA Xavier GPU. 
 

\section*{Acknowledgments}
This work was supported by the H2020 ITN project ACHIEVE (H2020-MSCA-ITN-2017: agreement no. 765866).

\bibliographystyle{model2-names}
\bibliography{refs}

\begin{thebibliography}{39}
\expandafter\ifx\csname natexlab\endcsname\relax\def\natexlab#1{#1}\fi
\providecommand{\url}[1]{\texttt{#1}}
\providecommand{\href}[2]{#2}
\providecommand{\path}[1]{#1}
\providecommand{\DOIprefix}{doi:}
\providecommand{\ArXivprefix}{arXiv:}
\providecommand{\URLprefix}{URL: }
\providecommand{\Pubmedprefix}{pmid:}
\providecommand{\doi}[1]{\href{http://dx.doi.org/#1}{\path{#1}}}
\providecommand{\Pubmed}[1]{\href{pmid:#1}{\path{#1}}}
\providecommand{\bibinfo}[2]{#2}
\ifx\xfnm\relax \def\xfnm[#1]{\unskip,\space#1}\fi
\bibitem[{Ali and Taylor(2018)}]{ali2018real}
\bibinfo{author}{Ali, A.}, \bibinfo{author}{Taylor, G.W.},
  \bibinfo{year}{2018}.
\newblock \bibinfo{title}{Real-time end-to-end action detection with two-stream
  networks}, in: \bibinfo{booktitle}{IEEE CRV}, pp. \bibinfo{pages}{31--38}.
\bibitem[{Brox et~al.(2004)Brox, Bruhn, Papenberg and Weickert}]{brox2004high}
\bibinfo{author}{Brox, T.}, \bibinfo{author}{Bruhn, A.},
  \bibinfo{author}{Papenberg, N.}, \bibinfo{author}{Weickert, J.},
  \bibinfo{year}{2004}.
\newblock \bibinfo{title}{High accuracy optical flow estimation based on a
  theory for warping}, in: \bibinfo{booktitle}{ECCV},
  \bibinfo{organization}{Springer}. pp. \bibinfo{pages}{25--36}.
\bibitem[{Carreira and Zisserman(2017)}]{carreira2017quo}
\bibinfo{author}{Carreira, J.}, \bibinfo{author}{Zisserman, A.},
  \bibinfo{year}{2017}.
\newblock \bibinfo{title}{Quo vadis, action recognition? a new model and the
  kinetics dataset}, in: \bibinfo{booktitle}{IEEE CVPR}, pp.
  \bibinfo{pages}{6299--6308}.
\bibitem[{Dai et~al.(2016)Dai, Li, He and Sun}]{dai2016r}
\bibinfo{author}{Dai, J.}, \bibinfo{author}{Li, Y.}, \bibinfo{author}{He, K.},
  \bibinfo{author}{Sun, J.}, \bibinfo{year}{2016}.
\newblock \bibinfo{title}{R-fcn: Object detection via region-based fully
  convolutional networks}, in: \bibinfo{booktitle}{NIPS}, pp.
  \bibinfo{pages}{379--387}.
\bibitem[{Dosovitskiy et~al.(2015)Dosovitskiy, Fischer, Ilg, Hausser, Hazirbas,
  Golkov, Van Der~Smagt, Cremers and Brox}]{dosovitskiy2015flownet}
\bibinfo{author}{Dosovitskiy, A.}, \bibinfo{author}{Fischer, P.},
  \bibinfo{author}{Ilg, E.}, \bibinfo{author}{Hausser, P.},
  \bibinfo{author}{Hazirbas, C.}, \bibinfo{author}{Golkov, V.},
  \bibinfo{author}{Van Der~Smagt, P.}, \bibinfo{author}{Cremers, D.},
  \bibinfo{author}{Brox, T.}, \bibinfo{year}{2015}.
\newblock \bibinfo{title}{Flownet: Learning optical flow with convolutional
  networks}, in: \bibinfo{booktitle}{IEEE ICCV}, pp.
  \bibinfo{pages}{2758--2766}.
\bibitem[{Feichtenhofer et~al.(2016)Feichtenhofer, Pinz and
  Zisserman}]{feichtenhofer2016convolutional}
\bibinfo{author}{Feichtenhofer, C.}, \bibinfo{author}{Pinz, A.},
  \bibinfo{author}{Zisserman, A.}, \bibinfo{year}{2016}.
\newblock \bibinfo{title}{Convolutional two-stream network fusion for video
  action recognition}, in: \bibinfo{booktitle}{IEEE CVPR}, pp.
  \bibinfo{pages}{1933--1941}.
\bibitem[{Girdhar et~al.(2019)Girdhar, Carreira, Doersch and
  Zisserman}]{girdhar2019video}
\bibinfo{author}{Girdhar, R.}, \bibinfo{author}{Carreira, J.},
  \bibinfo{author}{Doersch, C.}, \bibinfo{author}{Zisserman, A.},
  \bibinfo{year}{2019}.
\newblock \bibinfo{title}{Video action transformer network}, in:
  \bibinfo{booktitle}{IEEE CVPR}, pp. \bibinfo{pages}{244--253}.
\bibitem[{Gu et~al.(2018)Gu, Sun, Ross, Vondrick, Pantofaru, Li,
  Vijayanarasimhan, Toderici, Ricco, Sukthankar et~al.}]{gu2018ava}
\bibinfo{author}{Gu, C.}, \bibinfo{author}{Sun, C.}, \bibinfo{author}{Ross,
  D.A.}, \bibinfo{author}{Vondrick, C.}, \bibinfo{author}{Pantofaru, C.},
  \bibinfo{author}{Li, Y.}, \bibinfo{author}{Vijayanarasimhan, S.},
  \bibinfo{author}{Toderici, G.}, \bibinfo{author}{Ricco, S.},
  \bibinfo{author}{Sukthankar, R.}, et~al., \bibinfo{year}{2018}.
\newblock \bibinfo{title}{Ava: A video dataset of spatio-temporally localized
  atomic visual actions}, in: \bibinfo{booktitle}{IEEE CVPR}, pp.
  \bibinfo{pages}{6047--6056}.
\bibitem[{Han et~al.(2016)Han, Khorrami, Paine, Ramachandran, Babaeizadeh, Shi,
  Li, Yan and Huang}]{han2016seq}
\bibinfo{author}{Han, W.}, \bibinfo{author}{Khorrami, P.},
  \bibinfo{author}{Paine, T.L.}, \bibinfo{author}{Ramachandran, P.},
  \bibinfo{author}{Babaeizadeh, M.}, \bibinfo{author}{Shi, H.},
  \bibinfo{author}{Li, J.}, \bibinfo{author}{Yan, S.}, \bibinfo{author}{Huang,
  T.S.}, \bibinfo{year}{2016}.
\newblock \bibinfo{title}{Seq-nms for video object detection}.
\newblock \bibinfo{journal}{arXiv preprint arXiv:1602.08465} .
\bibitem[{Hetang et~al.(2017)Hetang, Qin, Liu and Yan}]{hetang2017impression}
\bibinfo{author}{Hetang, C.}, \bibinfo{author}{Qin, H.}, \bibinfo{author}{Liu,
  S.}, \bibinfo{author}{Yan, J.}, \bibinfo{year}{2017}.
\newblock \bibinfo{title}{Impression network for video object detection}.
\newblock \bibinfo{journal}{arXiv preprint arXiv:1712.05896} .
\bibitem[{Ilg et~al.(2017)Ilg, Mayer, Saikia, Keuper, Dosovitskiy and
  Brox}]{ilg2017flownet}
\bibinfo{author}{Ilg, E.}, \bibinfo{author}{Mayer, N.},
  \bibinfo{author}{Saikia, T.}, \bibinfo{author}{Keuper, M.},
  \bibinfo{author}{Dosovitskiy, A.}, \bibinfo{author}{Brox, T.},
  \bibinfo{year}{2017}.
\newblock \bibinfo{title}{Flownet 2.0: Evolution of optical flow estimation
  with deep networks}, in: \bibinfo{booktitle}{IEEE CVPR}, pp.
  \bibinfo{pages}{2462--2470}.
\bibitem[{Jhuang et~al.(2013)Jhuang, Gall, Zuffi, Schmid and
  Black}]{jhuang2013towards}
\bibinfo{author}{Jhuang, H.}, \bibinfo{author}{Gall, J.},
  \bibinfo{author}{Zuffi, S.}, \bibinfo{author}{Schmid, C.},
  \bibinfo{author}{Black, M.J.}, \bibinfo{year}{2013}.
\newblock \bibinfo{title}{Towards understanding action recognition}, in:
  \bibinfo{booktitle}{IEEE ICCV}, pp. \bibinfo{pages}{3192--3199}.
\bibitem[{Jiang et~al.(2019)Jiang, Wang, Gan, Wu and Yan}]{jiang2019stm}
\bibinfo{author}{Jiang, B.}, \bibinfo{author}{Wang, M.}, \bibinfo{author}{Gan,
  W.}, \bibinfo{author}{Wu, W.}, \bibinfo{author}{Yan, J.},
  \bibinfo{year}{2019}.
\newblock \bibinfo{title}{Stm: Spatiotemporal and motion encoding for action
  recognition}, in: \bibinfo{booktitle}{IEEE ICCV}, pp.
  \bibinfo{pages}{2000--2009}.
\bibitem[{Kalogeiton et~al.(2017)Kalogeiton, Weinzaepfel, Ferrari and
  Schmid}]{kalogeiton2017action}
\bibinfo{author}{Kalogeiton, V.}, \bibinfo{author}{Weinzaepfel, P.},
  \bibinfo{author}{Ferrari, V.}, \bibinfo{author}{Schmid, C.},
  \bibinfo{year}{2017}.
\newblock \bibinfo{title}{Action tubelet detector for spatio-temporal action
  localization}, in: \bibinfo{booktitle}{IEEE ICCV}, pp.
  \bibinfo{pages}{4405--4413}.
\bibitem[{Kang et~al.(2017)Kang, Li, Yan, Zeng, Yang, Xiao, Zhang, Wang, Wang,
  Wang et~al.}]{kang2017t}
\bibinfo{author}{Kang, K.}, \bibinfo{author}{Li, H.}, \bibinfo{author}{Yan,
  J.}, \bibinfo{author}{Zeng, X.}, \bibinfo{author}{Yang, B.},
  \bibinfo{author}{Xiao, T.}, \bibinfo{author}{Zhang, C.},
  \bibinfo{author}{Wang, Z.}, \bibinfo{author}{Wang, R.},
  \bibinfo{author}{Wang, X.}, et~al., \bibinfo{year}{2017}.
\newblock \bibinfo{title}{T-cnn: Tubelets with convolutional neural networks
  for object detection from videos}.
\newblock \bibinfo{journal}{IEEE TCSVT} \bibinfo{volume}{28},
  \bibinfo{pages}{2896--2907}.
\bibitem[{K{\"o}p{\"u}kl{\"u} et~al.(2019)K{\"o}p{\"u}kl{\"u}, Wei and
  Rigoll}]{kopuklu2019you}
\bibinfo{author}{K{\"o}p{\"u}kl{\"u}, O.}, \bibinfo{author}{Wei, X.},
  \bibinfo{author}{Rigoll, G.}, \bibinfo{year}{2019}.
\newblock \bibinfo{title}{You only watch once: A unified cnn architecture for
  real-time spatiotemporal action localization}.
\newblock \bibinfo{journal}{arXiv preprint arXiv:1911.06644} .
\bibitem[{Kroeger et~al.(2016)Kroeger, Timofte, Dai and
  Van~Gool}]{kroeger2016fast}
\bibinfo{author}{Kroeger, T.}, \bibinfo{author}{Timofte, R.},
  \bibinfo{author}{Dai, D.}, \bibinfo{author}{Van~Gool, L.},
  \bibinfo{year}{2016}.
\newblock \bibinfo{title}{Fast optical flow using dense inverse search}, in:
  \bibinfo{booktitle}{ECCV}, \bibinfo{organization}{Springer}. pp.
  \bibinfo{pages}{471--488}.
\bibitem[{Li et~al.(2019)Li, Miao, Tian, Fan, Xu, Ma and Song}]{li2019large}
\bibinfo{author}{Li, Y.}, \bibinfo{author}{Miao, Q.}, \bibinfo{author}{Tian,
  K.}, \bibinfo{author}{Fan, Y.}, \bibinfo{author}{Xu, X.},
  \bibinfo{author}{Ma, Z.}, \bibinfo{author}{Song, J.}, \bibinfo{year}{2019}.
\newblock \bibinfo{title}{Large-scale gesture recognition with a fusion of
  rgb-d data based on optical flow and the c3d model}.
\newblock \bibinfo{journal}{PRL} \bibinfo{volume}{119},
  \bibinfo{pages}{187--194}.
\bibitem[{Li et~al.(2020)Li, Wang, Wang and Wu}]{li2020actions}
\bibinfo{author}{Li, Y.}, \bibinfo{author}{Wang, Z.}, \bibinfo{author}{Wang,
  L.}, \bibinfo{author}{Wu, G.}, \bibinfo{year}{2020}.
\newblock \bibinfo{title}{Actions as moving points}.
\newblock \bibinfo{journal}{arXiv preprint arXiv:2001.04608} .
\bibitem[{Lin et~al.(2019)Lin, Gan and Han}]{lin2019tsm}
\bibinfo{author}{Lin, J.}, \bibinfo{author}{Gan, C.}, \bibinfo{author}{Han,
  S.}, \bibinfo{year}{2019}.
\newblock \bibinfo{title}{Tsm: Temporal shift module for efficient video
  understanding}, in: \bibinfo{booktitle}{IEEE ICCV}, pp.
  \bibinfo{pages}{7083--7093}.
\bibitem[{Liu and Zhu(2018)}]{liu2018mobile}
\bibinfo{author}{Liu, M.}, \bibinfo{author}{Zhu, M.}, \bibinfo{year}{2018}.
\newblock \bibinfo{title}{Mobile video object detection with temporally-aware
  feature maps}, in: \bibinfo{booktitle}{IEEE CVPR}, pp.
  \bibinfo{pages}{5686--5695}.
\bibitem[{Liu et~al.(2016)Liu, Anguelov, Erhan, Szegedy, Reed, Fu and
  Berg}]{liu2016ssd}
\bibinfo{author}{Liu, W.}, \bibinfo{author}{Anguelov, D.},
  \bibinfo{author}{Erhan, D.}, \bibinfo{author}{Szegedy, C.},
  \bibinfo{author}{Reed, S.}, \bibinfo{author}{Fu, C.Y.},
  \bibinfo{author}{Berg, A.C.}, \bibinfo{year}{2016}.
\newblock \bibinfo{title}{Ssd: Single shot multibox detector}, in:
  \bibinfo{booktitle}{ECCV}, \bibinfo{organization}{Springer}. pp.
  \bibinfo{pages}{21--37}.
\bibitem[{Peng and Schmid(2016)}]{peng2016multi}
\bibinfo{author}{Peng, X.}, \bibinfo{author}{Schmid, C.}, \bibinfo{year}{2016}.
\newblock \bibinfo{title}{Multi-region two-stream r-cnn for action detection},
  in: \bibinfo{booktitle}{ECCV}, \bibinfo{organization}{Springer}. pp.
  \bibinfo{pages}{744--759}.
\bibitem[{Qiu et~al.(2017)Qiu, Yao and Mei}]{qiu2017learning}
\bibinfo{author}{Qiu, Z.}, \bibinfo{author}{Yao, T.}, \bibinfo{author}{Mei,
  T.}, \bibinfo{year}{2017}.
\newblock \bibinfo{title}{Learning spatio-temporal representation with
  pseudo-3d residual networks}, in: \bibinfo{booktitle}{IEEE ICCV}, pp.
  \bibinfo{pages}{5533--5541}.
\bibitem[{Redmon and Farhadi(2017)}]{redmon2017yolo9000}
\bibinfo{author}{Redmon, J.}, \bibinfo{author}{Farhadi, A.},
  \bibinfo{year}{2017}.
\newblock \bibinfo{title}{Yolo9000: better, faster, stronger}, in:
  \bibinfo{booktitle}{IEEE CVPR}, pp. \bibinfo{pages}{7263--7271}.
\bibitem[{Ren et~al.(2015)Ren, He, Girshick and Sun}]{ren2015faster}
\bibinfo{author}{Ren, S.}, \bibinfo{author}{He, K.}, \bibinfo{author}{Girshick,
  R.}, \bibinfo{author}{Sun, J.}, \bibinfo{year}{2015}.
\newblock \bibinfo{title}{Faster r-cnn: Towards real-time object detection with
  region proposal networks}, in: \bibinfo{booktitle}{NIPS}, pp.
  \bibinfo{pages}{91--99}.
\bibitem[{Simonyan and Zisserman(2014)}]{simonyan2014two}
\bibinfo{author}{Simonyan, K.}, \bibinfo{author}{Zisserman, A.},
  \bibinfo{year}{2014}.
\newblock \bibinfo{title}{Two-stream convolutional networks for action
  recognition in videos}, in: \bibinfo{booktitle}{NIPS}, pp.
  \bibinfo{pages}{568--576}.
\bibitem[{Singh et~al.(2017)Singh, Saha, Sapienza, Torr and
  Cuzzolin}]{singh2017online}
\bibinfo{author}{Singh, G.}, \bibinfo{author}{Saha, S.},
  \bibinfo{author}{Sapienza, M.}, \bibinfo{author}{Torr, P.H.},
  \bibinfo{author}{Cuzzolin, F.}, \bibinfo{year}{2017}.
\newblock \bibinfo{title}{Online real-time multiple spatiotemporal action
  localisation and prediction}, in: \bibinfo{booktitle}{IEEE ICCV}, pp.
  \bibinfo{pages}{3637--3646}.
\bibitem[{Soomro et~al.(2012)Soomro, Zamir and Shah}]{soomro2012ucf101}
\bibinfo{author}{Soomro, K.}, \bibinfo{author}{Zamir, A.R.},
  \bibinfo{author}{Shah, M.}, \bibinfo{year}{2012}.
\newblock \bibinfo{title}{Ucf101: A dataset of 101 human actions classes from
  videos in the wild}.
\newblock \bibinfo{journal}{arXiv preprint arXiv:1212.0402} .
\bibitem[{Sun et~al.(2018a)Sun, Shrivastava, Vondrick, Murphy, Sukthankar and
  Schmid}]{sun2018actor}
\bibinfo{author}{Sun, C.}, \bibinfo{author}{Shrivastava, A.},
  \bibinfo{author}{Vondrick, C.}, \bibinfo{author}{Murphy, K.},
  \bibinfo{author}{Sukthankar, R.}, \bibinfo{author}{Schmid, C.},
  \bibinfo{year}{2018}a.
\newblock \bibinfo{title}{Actor-centric relation network}, in:
  \bibinfo{booktitle}{ECCV}, pp. \bibinfo{pages}{318--334}.
\bibitem[{Sun et~al.(2018b)Sun, Kuang, Sheng, Ouyang and
  Zhang}]{sun2018optical}
\bibinfo{author}{Sun, S.}, \bibinfo{author}{Kuang, Z.}, \bibinfo{author}{Sheng,
  L.}, \bibinfo{author}{Ouyang, W.}, \bibinfo{author}{Zhang, W.},
  \bibinfo{year}{2018}b.
\newblock \bibinfo{title}{Optical flow guided feature: A fast and robust motion
  representation for video action recognition}, in: \bibinfo{booktitle}{IEEE
  CVPR}, pp. \bibinfo{pages}{1390--1399}.
\bibitem[{Tran et~al.(2018)Tran, Wang, Torresani, Ray, LeCun and
  Paluri}]{tran2018closer}
\bibinfo{author}{Tran, D.}, \bibinfo{author}{Wang, H.},
  \bibinfo{author}{Torresani, L.}, \bibinfo{author}{Ray, J.},
  \bibinfo{author}{LeCun, Y.}, \bibinfo{author}{Paluri, M.},
  \bibinfo{year}{2018}.
\newblock \bibinfo{title}{A closer look at spatiotemporal convolutions for
  action recognition}, in: \bibinfo{booktitle}{IEEE CVPR}, pp.
  \bibinfo{pages}{6450--6459}.
\bibitem[{Wang et~al.(2016)Wang, Xiong, Wang, Qiao, Lin, Tang and
  Van~Gool}]{wang2016temporal}
\bibinfo{author}{Wang, L.}, \bibinfo{author}{Xiong, Y.}, \bibinfo{author}{Wang,
  Z.}, \bibinfo{author}{Qiao, Y.}, \bibinfo{author}{Lin, D.},
  \bibinfo{author}{Tang, X.}, \bibinfo{author}{Van~Gool, L.},
  \bibinfo{year}{2016}.
\newblock \bibinfo{title}{Temporal segment networks: Towards good practices for
  deep action recognition}, in: \bibinfo{booktitle}{ECCV},
  \bibinfo{organization}{Springer}. pp. \bibinfo{pages}{20--36}.
\bibitem[{Wei et~al.(2019)Wei, Wang, Yi, Li and Huang}]{wei2019p3d}
\bibinfo{author}{Wei, J.}, \bibinfo{author}{Wang, H.}, \bibinfo{author}{Yi,
  Y.}, \bibinfo{author}{Li, Q.}, \bibinfo{author}{Huang, D.},
  \bibinfo{year}{2019}.
\newblock \bibinfo{title}{P3d-ctn: Pseudo-3d convolutional tube network for
  spatio-temporal action detection in videos}, in: \bibinfo{booktitle}{IEEE
  ICIP}, pp. \bibinfo{pages}{300--304}.
\bibitem[{Yang et~al.(2019)Yang, Yang, Liu, Xiao, Davis and
  Kautz}]{yang2019step}
\bibinfo{author}{Yang, X.}, \bibinfo{author}{Yang, X.}, \bibinfo{author}{Liu,
  M.Y.}, \bibinfo{author}{Xiao, F.}, \bibinfo{author}{Davis, L.S.},
  \bibinfo{author}{Kautz, J.}, \bibinfo{year}{2019}.
\newblock \bibinfo{title}{Step: Spatio-temporal progressive learning for video
  action detection}, in: \bibinfo{booktitle}{IEEE CVPR}, pp.
  \bibinfo{pages}{264--272}.
\bibitem[{Yao et~al.(2019)Yao, Lei and Zhong}]{yao2019review}
\bibinfo{author}{Yao, G.}, \bibinfo{author}{Lei, T.}, \bibinfo{author}{Zhong,
  J.}, \bibinfo{year}{2019}.
\newblock \bibinfo{title}{A review of convolutional-neural-network-based action
  recognition}.
\newblock \bibinfo{journal}{PRL} \bibinfo{volume}{118},
  \bibinfo{pages}{14--22}.
\bibitem[{Zhao and Snoek(2019)}]{zhao2019dance}
\bibinfo{author}{Zhao, J.}, \bibinfo{author}{Snoek, C.G.},
  \bibinfo{year}{2019}.
\newblock \bibinfo{title}{Dance with flow: Two-in-one stream action detection},
  in: \bibinfo{booktitle}{IEEE CVPR}, pp. \bibinfo{pages}{9935--9944}.
\bibitem[{Zhu et~al.(2017a)Zhu, Wang, Dai, Yuan and Wei}]{zhu2017flow}
\bibinfo{author}{Zhu, X.}, \bibinfo{author}{Wang, Y.}, \bibinfo{author}{Dai,
  J.}, \bibinfo{author}{Yuan, L.}, \bibinfo{author}{Wei, Y.},
  \bibinfo{year}{2017}a.
\newblock \bibinfo{title}{Flow-guided feature aggregation for video object
  detection}, in: \bibinfo{booktitle}{IEEE ICCV}, pp.
  \bibinfo{pages}{408--417}.
\bibitem[{Zhu et~al.(2017b)Zhu, Xiong, Dai, Yuan and Wei}]{zhu2017deep}
\bibinfo{author}{Zhu, X.}, \bibinfo{author}{Xiong, Y.}, \bibinfo{author}{Dai,
  J.}, \bibinfo{author}{Yuan, L.}, \bibinfo{author}{Wei, Y.},
  \bibinfo{year}{2017}b.
\newblock \bibinfo{title}{Deep feature flow for video recognition}, in:
  \bibinfo{booktitle}{IEEE CVPR}, pp. \bibinfo{pages}{2349--2358}.

\end{thebibliography}

\end{document}